\documentclass{article} 
\usepackage{RobustML_iclr2021_conference,times}

\usepackage{amsmath,amsfonts,bm}

\def\eqref#1{equation~\ref{#1}}

\def\1{\bm{1}}

\DeclareMathAlphabet{\mathsfit}{\encodingdefault}{\sfdefault}{m}{sl}
\SetMathAlphabet{\mathsfit}{bold}{\encodingdefault}{\sfdefault}{bx}{n}

\usepackage{amsmath,graphicx,amsfonts,amssymb}
\usepackage{hyperref}
\hypersetup{colorlinks,linkcolor={blue},citecolor={blue},urlcolor={blue}}
\usepackage{url}
\usepackage{bm}
\usepackage{graphicx}

\usepackage{listings}
\usepackage{color}

\definecolor{dkgreen}{rgb}{0,0.6,0}
\definecolor{gray}{rgb}{0.5,0.5,0.5}
\definecolor{mauve}{rgb}{0.58,0,0.82}

\usepackage[compact]{titlesec}

\title{Tasting the cake: evaluating self-supervised generalization on out-of-distribution multimodal MRI data}

\iclrfinalcopy

\author{Alex Fedorov \\
Georgia Institute of Technology \\
TReNDS Center, Atlanta, GA, USA \\
\texttt{afedorov@gatech.edu}
\And
Eloy Geenjaar \\
Delft University of Technology \\
Delft, the Netherlands \\
TReNDS Center, Atlanta, GA, USA
\And
Lei Wu,
Thomas P. DeRamus,
Vince D. Calhoun,
Sergey M. Plis \\
TReNDS Center, Atlanta, GA, USA
}

\begin{document}

\maketitle

\begin{abstract}
  Self-supervised learning has enabled significant improvements on natural image benchmarks. However, there is less work in the medical imaging domain in this area. The optimal models have not yet been determined among the various options.
  Moreover, little work has evaluated the current applicability limits of novel self-supervised methods.
  In this paper, we evaluate a range of current contrastive self-supervised methods on out-of-distribution generalization in order to evaluate their applicability to medical imaging.
  We show that self-supervised models are not as robust as expected based on their results in natural imaging benchmarks and can be outperformed by supervised learning with dropout.
  We also show that this behavior can be countered with extensive augmentation.
  Our results highlight the need for out-of-distribution generalization standards and benchmarks to adopt the self-supervised methods in the medical imaging community.
  \end{abstract}
\vspace{-0.25cm}

\section{Introduction}

Self-supervised learning has fueled recent advances in image recognition~\citep{oord2018representation, hjelm2018learning, bachman2019learning, tian2019contrastive, chen2020simple, grill2020bootstrap, chen2020exploring} and spurred great interest and high expectations in neuroimaging~\citep{fedorov2019prediction, mahmood2020whole, jeon2020enriched, NEURIPS2020_d2dc6368, fedorov2020self}.
The expectations are generally high even outside neuroimaging, so much so that in Yann LeCun's metaphor of learning as a cake~\citep{YannLeCunCake}, self-supervised learning makes the tastiest part of the cake: the filling.

However, the observed similarity in performance of self-supervised and supervised methods on natural images~\citep{geirhos2020surprising} does not guarantee the same for other domains. Work on the generalization of these methods in neuroimaging data is generally lacking.
We investigate the out-of-distribution generalization using simulated distortions and a natural distributional shift based on race with multimodal human MRI data to fill this gap.
The multimodal data is as a natural case of multi-view data. It contains a wealth of complementary information regarding the healthy and dysfunctional brain~\citep{calhoun2016multimodal}.
We show that on neuroimaging data, contrastive multimodal self-supervised learning leads to models that differ from models trained in a supervised way.\footnote{Further, for brevity we use the terms supervised models and self-supervised models, but what we mean is the models trained by these approaches to learning.}
The models disagree in how they react to distortions or modality.
When using Dropout~\citep{tompson2015efficient}, supervised models, counter to our expectations, can significantly outperform self-supervised models in out-of-distribution generalization.
Further, the class of methods inspired by DeepInfoMax (DIM)~\citep{hjelm2018learning} tends to struggle with intensity-based distortions, which we attempt to solve using additional data augmentation.
Our findings further reinforce the advantages of multimodal models over unimodal ones.

Finally, we argue that the medical imaging community needs
standards and benchmarks for out-of-distribution generalization.
Introducing them could advance researchers towards a more reliable and standardized
evaluation of newly proposed methods because similar performance on
downstream tasks does not necessarily imply robust
generalization.
Standards could lead to a better understanding of
methodological trade-offs in medical imaging. For example,
the out-of-distribution generalization to distortions (e.g., affine scale
distortion) may drive models to learn trivial discriminative features
(see shortcut~\citep{geirhos2020shortcut}).

\section{Methodology}

\subsection{Dataset and modeling out-of-distribution generalization}
We evaluate the model on multimodal neuroimaging dataset OASIS-3~\citep{LaMontagne2019.12.13.19014902}. The modalities we selected are T1 and resting-state fMRI (rs-fMRI), which capture the brain's anatomy and functional dynamics, respectively. First, T1 volumes were masked to only include the brain, and the rs-fMRI volumes were used to compute the fractional amplitude of low-frequency fluctuation (fALFF) in the $0.01$ to $0.1$Hz power band. The exhaustive details can be found in Appendix~\ref{appendix:preprocessing}.

To model out-of-distribution data, we utilize the following random data transformations available in TorchIO~\citep{perez-garcia_torchio_2020}: Affine, Elastic, Gamma, Motion, Spike, Ghosting, BiasField, GaussianNoise. The transformation details can be found in Appendix~\ref{appendix:distortions}.
The samples are shown in the Appendix in Figure~\ref{fig:samples}.
To model natural distributional shift from the training set, we selected African-American subjects.

\begin{figure}[t]
  \vspace{-1.4cm}
  \centering
  \begin{minipage}[b]{0.3\linewidth}
    \includegraphics[width=\linewidth]{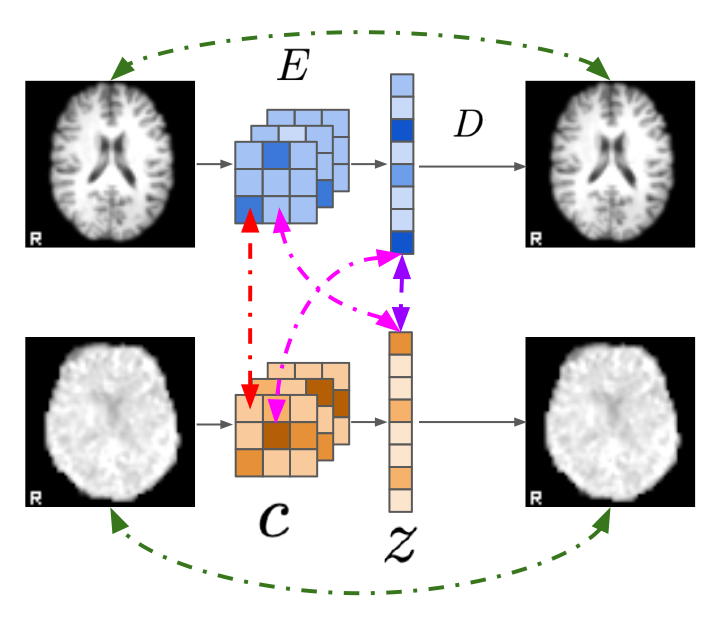}
  \end{minipage}
  \hfill
  \begin{minipage}[b]{0.5\linewidth}
    \includegraphics[width=\linewidth]{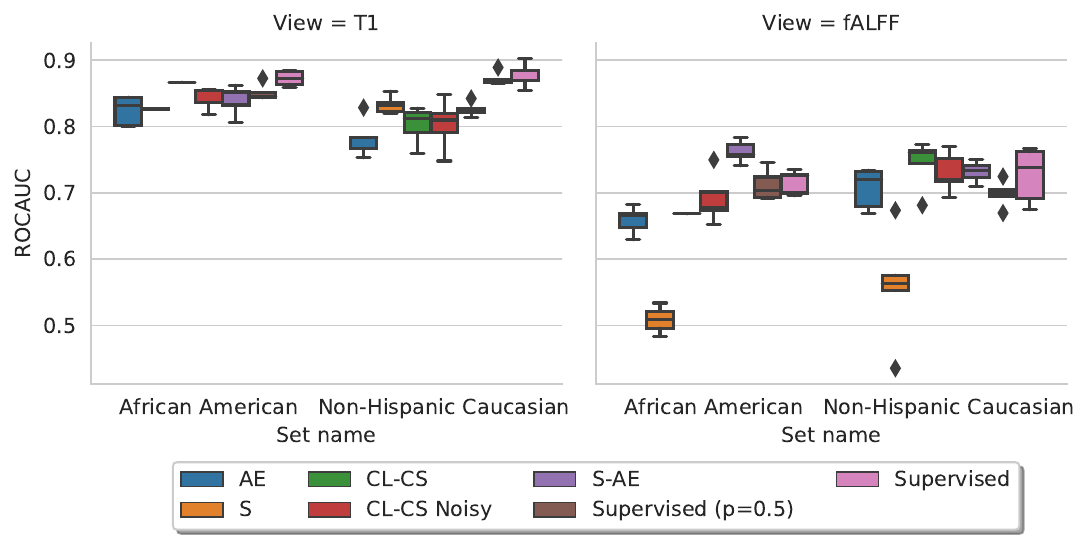}
  \end{minipage}
  \caption{Left: The combined scheme of CL--CS and S--AE. Red arrow defines CS, Pink --- CL, Purple --- S, and Green --- reconstruction for AE. Decoder $D$ is only required for S--AE, as CL--CS is a decoder-free model. Right: ROCAUC for holdout test sets with Non-Hispanic Caucausian and African American subjects for T1 and fALFF.}
  \label{fig:performance}
  \vspace{-0.5cm}
\end{figure}

\subsection{Multimodal self-supervised models}

Let ${\cal D} = \{x^{\text{T1}}_i, x^{\text{fALFF}}_i\}_{i=1}^N$ be a dataset with $N$ paired T1 and fALFF volumes. We want to learn a latent representation $z^M = E^M_{\theta_M}(x^M) \in \mathbb{R}^{64}$ where $E^m_{\theta_m}$ is an encoder part of the 3D convolutional DCGAN architecture~\citep{radford2015unsupervised} for modality $M$.

Based on the performance of the proposed taxonomy in~\cite{fedorov2020taxonomy}, we selected the following multimodal self-supervised models: CL--CS (Cross-Local and Cross-Spatial connections) and S--AE (Similarity AutoEncoder) (Figure~\ref{fig:performance}). Additionally, we compare these models with Supervised (pretraining with cross-entropy) and AutoEncoder (AE) unimodal models, and multimodal models based only on the maximization of similarity (S) between latent representations. The model S is equivalent to SimCLR~\citep{chen2020simple}.

CL--CS is an AMDIM~\citep{bachman2019learning} inspired model. Its inductive bias is to maximize mutual information between a p`air of global and local variables. The first part of the objective, called CL, is defined between a "local" variable $c$ (embedding of the location in the convolutional featuremap) in layer $l$ and a "global" (latent) variable $z$ as $\{(c_{j,l}^M, z^L)_{M \ne L}\}$, where $j$ is the location index, and $M$ and $L$ are modality indexes. The second part of the objective, called CS, is defined on pairs $\{(c_{j,l}^M, c_{k,l}^L)_{M \ne L}\}$ where $k$ is a location index in the other modality. The CL--CS objective is by minimizing ${\cal L}_{\text{CL--CS}} = I_{\text{CL}}(\{c\}_{\text{T1}}; \{z\}_{\text{fALFF}}) + I_{\text{CL}}(\{c\}_{\text{fALFF}}; \{z\}_{\text{T1}}) + I_{\text{CS}}(\{c\}_{\text{T1}}; \{c\}_{\text{fALFF}}) + I_{\text{CS}}(\{c\}_{\text{fALFF}}; \{c\}_{\text{T1}})$,
where $I(\{u\}_M; \{v\}_L) = -\frac{1}{N} \sum_{n=1}^N \log \frac{e^{f(u^M_n,v^L_n)}}{\frac{1}{N} \sum_{k=1}^N e^{f(u_n^M,v_k^L)}}$ is an InfoNCE~\citep{oord2018representation} based estimator with a separable critic $f(u^M_n, v^L_k) = \frac{u^{M\intercal}_n v^L_k}{\sqrt{n}}$~\citep{bachman2019learning} and $n =64$ is the dimensionality of the embeddings.

S--AE is a fusion of  CMC~\citep{tian2019contrastive} and SimCLR~\citep{chen2020simple} with DCCAE~\citep{wang2015deep}, where the CCA objective is substituted by a maximization of similarity (mutual information) between a pair of latent variables $\{(z^M, z^L)_{M \ne L}\}$. The similarity is approximated with a DIM objective that allows the model to be trained end-to-end and improves numerical stability compared to SVD-based solutions of CCA~\citep{wang2015deep}. Combining an AE with similarity maximization is shown to avoid the collapse of its representation in a multimodal scenario~\citep{fedorov2020taxonomy}. The S--AE objective is by minimizing $
    {\cal L}_{\text{S--AE}} = I(\{z\}_{\text{T1}}; \{z\}_{\text{fALFF}}) + I(\{z\}_{\text{fALFF}}; \{z\}_{\text{T1}}) + R^{\text{T1}} + R^{\text{fALFF}}
$, where $R^M = \frac{1}{N}  \sum_{n=1}^N ||x^M_n - D^M(E^M(x^M_n))||^2$ is a reconstruction loss with a DCGAN decoder $D$ for modality $M$.
\vspace{-0.25cm}
\section{Results and Discussion}

In our experiments, we use the frozen encoder and linear projection trained on Non-Hispanic Caucasian subjects. First, we pretrain the encoder on all possible pairs; then, we train a linear projection from the encoder's output to classify Healthy Controls (HC) and patients with Alzheimer's Disease (AD). Eventually, we evaluate the out-of-distribution generalization on a natural race-based distributional shift and simulated distortions applied to volumes in standardized MNI space. All the models follow the same pipeline and use the same hyperparameters (Appendix~\ref{appendix:hyperparameters}).

\begin{figure}[t]
  \vspace{-1.4cm}
  \centering
  \includegraphics[width=0.85\linewidth]{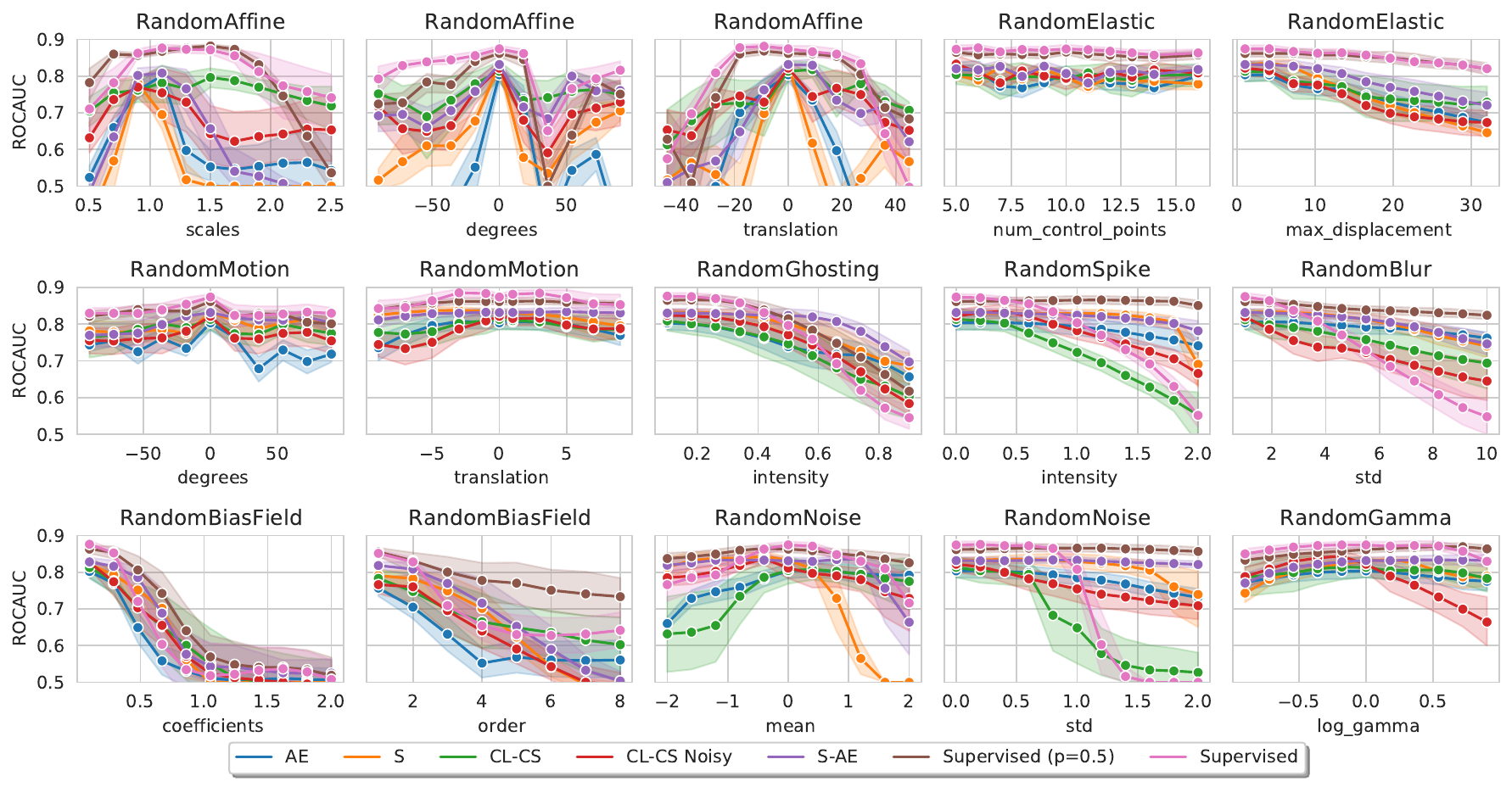}
  \caption{ROCAUC performance for out-of-distribution generalization on T1 data with simulated distortions. Each title contains the name of a random distorsion and x-axis label --- parameter of the distorsion.}
  \label{fig:results_t1}
  \vspace{-0.5cm}
\end{figure}

\subsection{Do self-supervised multimodal models produce robust representations?}

In Figure~\ref{fig:results_t1} we compare the results of selected models and their baselines on out-of-distributional generalization with simulated distortions. In most cases, the supervised model performs quite well and better than the self-supervised models, except it may break down on RandomGhosting, RandomSpike, RandomBlur, and Random Noise for stronger distortion levels. The unimodal AE and multimodal S perform much worse compared to the combined approach: S--AE. CL--CS  performs much worse than other models on intensity-based noise: RandomGhosting, RandomSpike, RandomBlur, RandomNoise. We hypothesize that adding intensity-based data-augmentation should help DIM-inspired models. DIM-inspired models maximize the mutual information between features with respect to depth, which can lead to learning spurious correlations.

\begin{figure}[t]
  \vspace{-1.4cm}
  \centering
  \includegraphics[width=0.85\linewidth]{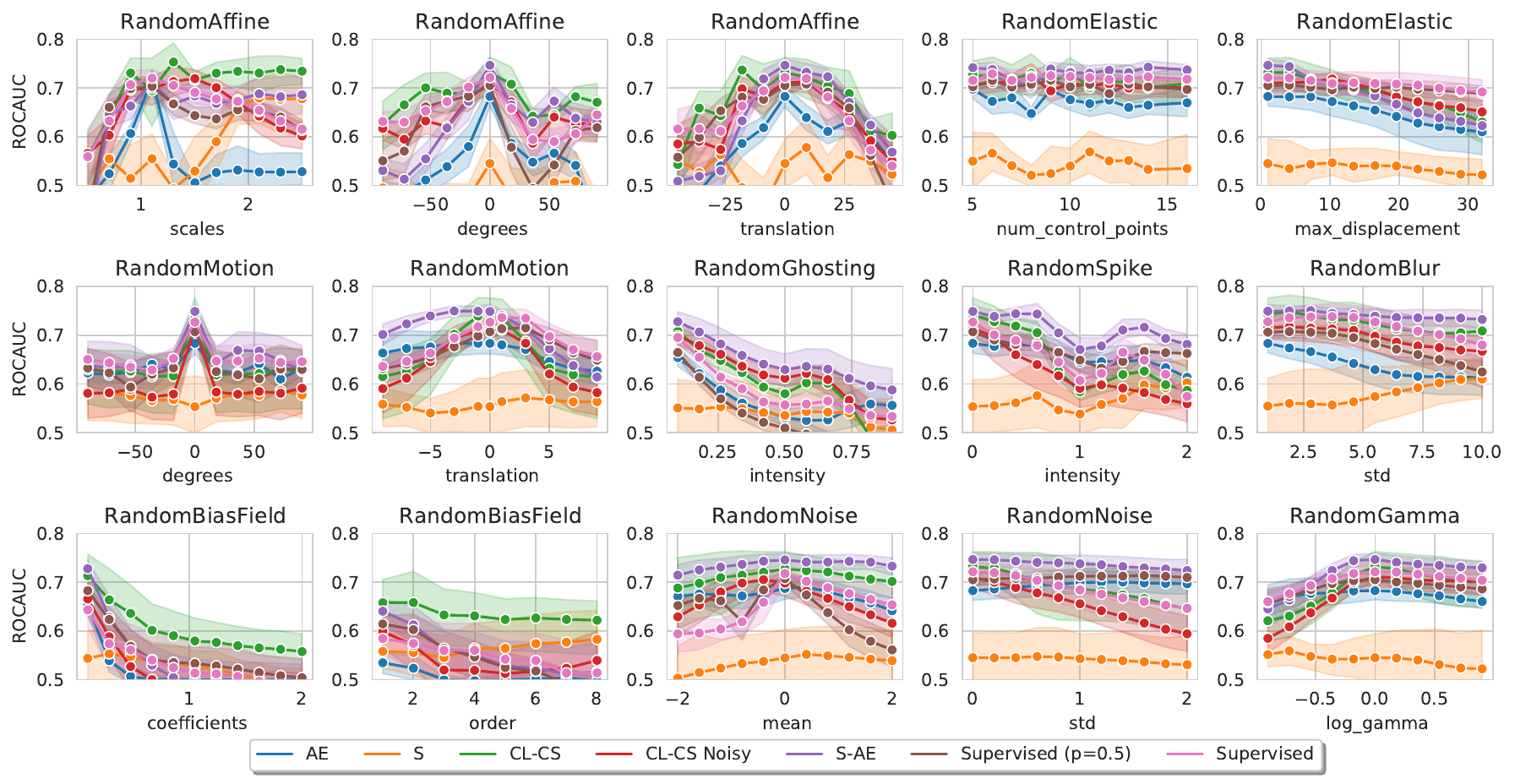}
  \caption{ROCAUC performance for out-of-distribution generalization on fALFF data with simulated distortions. Each title contains the name of a random distorsion and x-axis label --- parameter of the distorsion.}
  \label{fig:results_falff}
  \vspace{-0.5cm}
\end{figure}

In Figure~\ref{fig:results_falff} we compare models on out-of-distributional performance for fALFF. Specifically, in this case, S completely fails to represent fALFF, and the unimodal AE may fail in most cases. The multimodal self-supervised models: S--AE and CL--CS outperform the supervised baseline in most cases.

We want to note that fALFF is a "harder" modality because it represents rs-fMRI as a timeless and less-informative voxel-wise hand-engineered feature. Visually (Figure~\ref{fig:performance} (left)), it looks highly noisy compared to T1. Importantly, most of the distortions used to augment the data are not natural variations in the fALFF data because it has undergone a heavy preprocessing pipeline.

It is unclear what generalization requirements we should satisfy in medical imaging and what out-of-distribution generalization is meaningful. For example, if we look at the first subfigure with the RandomAffine and scale parameter in Figure~\ref{fig:results_t1}, it is not clear whether generalization to scaling is desirable. When we scale the volume 1.7$\times$ or more, we can only see the center of the brain, which might force some models (Supervised, CL--CS) to learn trivial features, such as a reduced ventricle size is a well-known trivial biomarker for Alzheimer's disease~\citep{frisoni2010clinical}. In contrast, other models (AE, S, S--AE) try to utilize information from the whole brain for decision making because they aren't directly optimizing an objective that maximizes classification accuracy.

These findings suggest contrastive multimodal self-supervised learning produces models that differ strongly from models trained in a supervised way but that there is room for improvement for both methods. We explore the improvement of Supervised and DIM models in the next subsection.

\subsubsection{Improving out-of-distribution generalization of supervised and DIM inspired models}

To improve the supervised model's out-of-distribution generalization, we utilized volumetric (3D) Dropout with $p=0.5$. The Supervised model with dropout shows a significant boost in generalization for T1 volumes compared to self-supervised models (Supervised (p=0.5), Figure~\ref{fig:results_t1}). Additionally, dropout did push the model to not only extract features from the center of the brain Figure~\ref{fig:results_t1} for RandomAffine scale distortion. This may suggest that dropout is a simple solution to the problem. Dropout however, does not work for noisy and hand-engineered fALFF data (Figure~\ref{fig:results_falff}).

To improve the out-of-distribution generalization for CL--CS, we added RandomNoise with a maximum standard deviation of $1.0$ with a probability of $0.33$ to the data augmentation pipeline during the encoder's pretraining. Such data augmentations partly improve generalization for T1 (CL--CS Noisy, Figure~\ref{fig:results_t1}) but may reduce out-of-distribution generalization for some distortions and does not work for fALFF. This solution requires some additional finetuning. Some ideas to improve data-augmentation are to utilize curriculum learning (e.g.,~\cite{NEURIPS2020_f6a673f0}) due convergence improvements.

\subsection{Race-based distributional shift}

The performance of the selected models is shown in Figure~\ref{fig:performance} (right). When comparing the models' performance on different races visually, there is no evident racial bias trend. The  S model fails to generalize to African American subjects with a classification accuracy of $0.5$ for fALFF. The reduced classification accuracy is likely due to its representations collapsing during pretraining.

\section{Conclusions}
Self-supervised medical imaging models are only now beginning to be developed, and we hope that our analysis will facilitate robust and fair self-supervised models. Additionally, we hope to see more exhaustive benchmarks to evaluate out-of-distribution generalization in medical imaging.

\subsubsection*{Acknowledgments}

This work is supported by NIH R01 EB006841.

Data were provided in part by OASIS-3: Principal Investigators: T. Benzinger, D. Marcus, J. Morris; NIH P50 AG00561, P30 NS09857781, P01 AG026276, P01 AG003991, R01 AG043434, UL1 TR000448, R01 EB009352. AV-45 doses were provided by Avid Radiopharmaceuticals, a wholly-owned subsidiary of Eli Lilly.

\bibliography{references}
\bibliographystyle{iclr2021_conference}

\newpage
\appendix
\section{Appendix}
\subsection{Simulated distortions}
\label{appendix:distortions}
\begin{figure}[!ht]
  \centering
  \includegraphics[width=\linewidth]{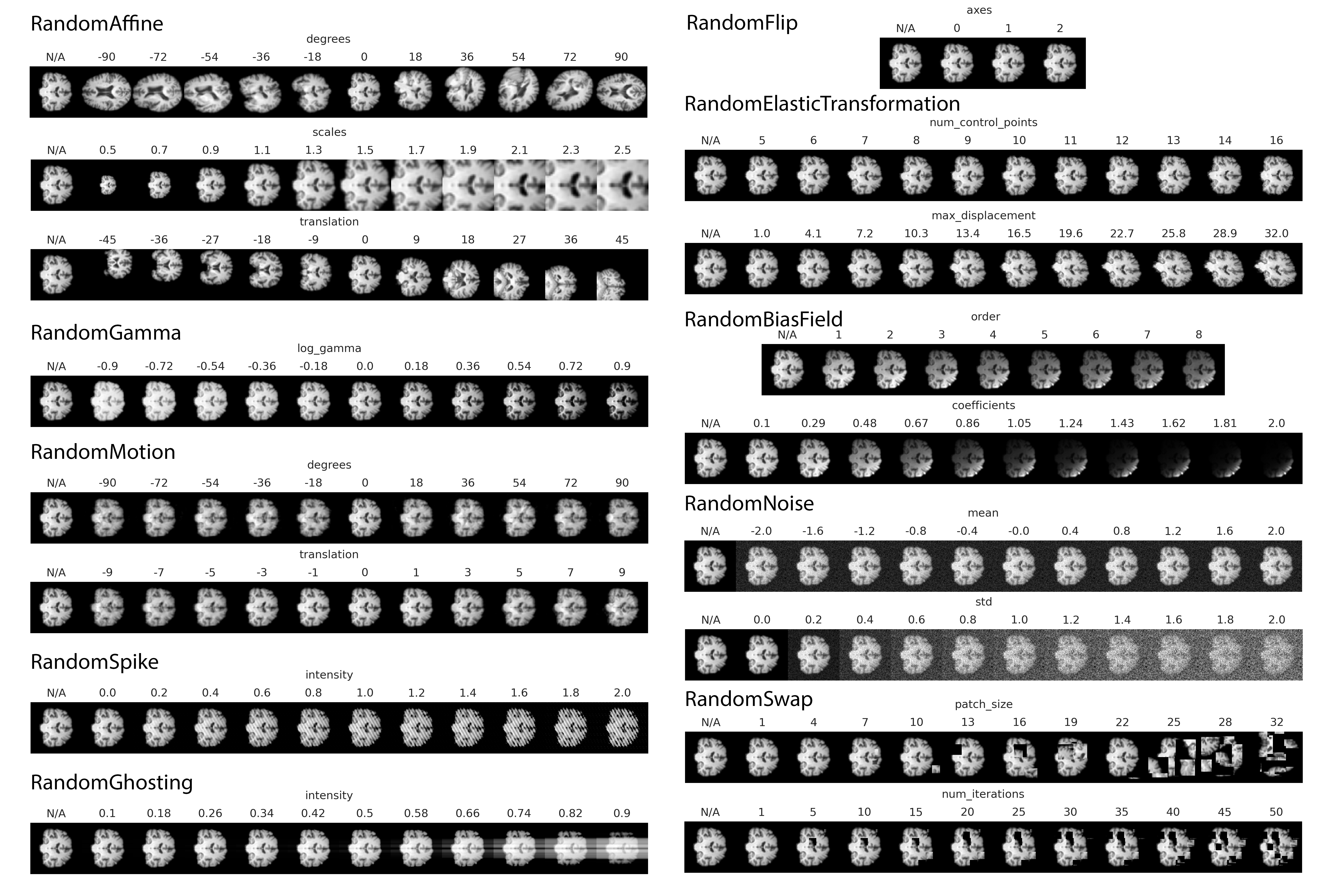}
  \caption{The simulated samples are shown on T1 image.}
  \label{fig:samples}
\end{figure}

The default parameters for data transformation are:
\begin{lstlisting}
"affine": {
  "scales": (1, 1), "degrees": (0, 0), "translation": (0, 0)
}
"anisotropy": {"downsampling": (2, 2)}
"motion": {
  "degrees": (0, 0), "translation": (0, 0), "num_transforms": 1
}
"ghost": {"num_ghosts": (1, 1), "intensity": (0.5, 0.5)}
"spike": {"num_spikes": (1, 1), "intensity": (0.5, 0.5)}
"blur": {"std": (0.25, 0.25)}
"bias": {"coefficients": (0.5, 0.5), "order": 3}
"noise": {"mean": (0, 0), "std": (0.25, 0.25)}
\end{lstlisting}
For transformations that are not listed, we have used default parameters from TorchIO package~\citep{perez-garcia_torchio_2020}.

The parameter space is defined as:
\begin{lstlisting}
class ParameterSpace():
  def __init__(self, left, right, n_steps, data_type, eps=10e-8):
      step = (right - left) / n_steps
      self.range = np.round(np.arange(left, right + eps, step), 2)
      self.range = self.range.astype(data_type)
\end{lstlisting}
Then we can define search space as:
\begin{lstlisting}
search_space = {
  "affine": {
      "class": tio.transforms.RandomAffine,
      "space": {
          "scales": ParameterSpace(0.5, 2.5, 10, data_type=float),
          "degrees": ParameterSpace(-90, 90, 10, data_type=int),
          "translation": ParameterSpace(-45, 45, 10, data_type=int)
      }
  },
  "elastic": {
      "class": tio.transforms.RandomElasticDeformation,
      "space": {
          "num_control_points": ParameterSpace(
              5, 16, 10, data_type=int),
          "max_displacement": ParameterSpace(
              1, 32, 10, data_type=float),
      }
  },
  "motion": {
      "class": tio.transforms.RandomMotion,
      "space": {
          "degrees": ParameterSpace(-90, 90, 10, data_type=int),
          "translation": ParameterSpace(-9, 9, 10, data_type=int),
      }
  },
  "ghost": {
      "class": tio.transforms.RandomGhosting,
      "space": {
          "intensity": ParameterSpace(0.1, 0.9, 10, data_type=float),
      }
  },
  "spike": {
      "class": tio.transforms.RandomSpike,
      "space": {
          "intensity": ParameterSpace(0, 2, 10, data_type=float)
      }
  },
  "blur": {
      "class": tio.transforms.RandomBlur,
      "space": {
          "std": ParameterSpace(1, 10, 10, data_type=float)
      }
  },
  "bias": {
      "class": tio.transforms.RandomBiasField,
      "space": {
          "coefficients": ParameterSpace(
              0.1, 2, 10, data_type=float),
          "order": ParameterSpace(1, 8, 8, data_type=int)
      }
  },
  "noise": {
      "class": tio.transforms.RandomNoise,
      "space": {
          "mean": ParameterSpace(-2., 2., 10, data_type=float),
          "std": ParameterSpace(0, 2., 10, data_type=float)
      }
  },
  "gamma": {
      "class": tio.transforms.RandomGamma,
      "space": {
          "log_gamma": ParameterSpace(-0.9, 0.9, 10, data_type=float)
      }
  }
}
\end{lstlisting}

The data transformation class is defined by a class in the TorchIO package~\citep{perez-garcia_torchio_2020}. The sample images for each transformation are shown in Figure~\ref{fig:samples}.

\subsection{Dataset preprocessing details}
\label{appendix:preprocessing}
To preprocess rs-fMRI, we registered the time-series to the first image in the series using mcflirt (FSL v6.0.2)~\citep{fsl} with a 3-stage search level ($8$mm, $4$mm, $4$mm), 20mm field-of-view, matching with 256 histogram bins, 6 degrees-of-freedom (dof) for the transformation, a 6mm scaling factor, and normalized correlation values across the volumes as a cost function (smoothed to 1mm). The interpolation of the final transformations and outputs is done using splines. The fALFF volume was then computed in the 0.01 to 0.1 Hz power band using REST~\citep{rest}.

To preprocess the T1 volume, we first removed 15 subjects after visual inspection. The T1 volumes were brainmasked with bert ((FSL v6.0.2)~\citep{fsl}). The brainmasked volumes were then linearly warped (7 dof) to MNI space and resampled to a 3mm resolution with a final volume of 64-by-64-by-64 mm.

The samples were z-normalized and normalized with histogram standardization based on the training set before being fed into the deep neural network. During training, we apply random flips and random crops as data augmentation.

We selected $826$ non-Hispanic Caucasian subjects ($70\%$ HC, $15\%$ AD, $15\%$ with other disorders). After pairing the two modalities for all of the scans the dataset included a total of $4021$ pairs, because subjects can have multiple scans. We split the subjects into $5$ stratified folds ($580-582$ subjects ($2828-2944$ pairs), $144-146$ ($653-769$)) and hold-out --- $100$ ($424$). The subset with African American subjects contains $134$ ($100$ HC, $34$ AD) samples. In the downstream tasks we only use the first pair of multimodal volumes per subject.

\subsection{Architecture, optimization and hyperparameters}
\label{appendix:hyperparameters}
The main architectures for the encoder and decoder are based on the fully convolutional DCGAN architecture ~\citep{radford2015unsupervised}. The final convolutional layer in the encoder produces a 64x1x1x1 feature map. We initialize all layers with Xavier initialization.

The CL--CS method also uses a convolutional projection head to map $c$ (128x8x8x8) to a 64x8x8x8 to get 8x8x8 locations with 64-dimensional representation. The projection head consists of one ResNet~\citep{he2016deep} block, which combines information from two paths: identity and two convolutional layers with a kernel size of $1$, $64$ channels, ReLU activation function and 3D Batch Normalization in between. The projections are shared between the CL and CS objectives. The layers of the convolutional projection are initialized as a uniform distribution $[-0.01, 0.01]$ and set to $1.$ on the diagonal, which is where the input and output dimensions match, similar to AMDIM~\citep{bachman2019learning}.

Similarly to AMDIM~\citep{bachman2019learning}, each InfoNCE objective is penalized using squared critic scores $\lambda f(u,v)^2$ where $\lambda = 4\mathrm{e}{-2}$ and we clip the values of the critic by $c\tanh(\frac{s}{c})$ with $c=20$.

We pretrain the encoders for $200$ epochs and train the linear projection layers for $500$ epochs with batch size $64$ using the RAdam~\citep{liu2019variance} optimizer (learning rate $4e-4$) and a OneCycleLR~\citep{smith2019super} scheduler (maximum learning rate $0.01$).

\subsection{Implementation and computational resources}

The experiments are implemented using PyTorch~\citep{paszke2019pytorch} and Catalyst~\citep{catalyst} frameworks. The experiments were performed with NVIDIA DGX-1 V100. The code will be available at some point.

\end{document}